\documentclass{template/llncs}
\usepackage[T1]{fontenc}
\usepackage{graphicx}
\usepackage{orcidlink}
\usepackage{bbding} 
\usepackage{enumitem} 
\usepackage{multirow}
\usepackage{multicol}
\usepackage{makecell}
\usepackage{amssymb}
\usepackage{listings}
\usepackage{xcolor}
\usepackage{amsmath}
\usepackage{pgf-umlsd}
\usepackage{tikz}
\usetikzlibrary{positioning,arrows.meta}
\usepackage{subcaption}
\usepackage{hyperref}

\definecolor{jsonkey}{rgb}{0.0,0.3,0.6}
\definecolor{jsonstring}{rgb}{0.6,0.1,0.1}
\definecolor{jsonbg}{rgb}{1, 1, 1}
\definecolor{jsonnum}{rgb}{0.1,0.5,0.1}

\lstdefinestyle{jsonstyle}{
backgroundcolor=\color{jsonbg},
basicstyle=\ttfamily\footnotesize,
breaklines=true,
showstringspaces=false,
numbers=left,
numberstyle=\tiny\color{gray},
stringstyle=\color{jsonstring},
keywordstyle=\color{jsonkey},
commentstyle=\color{gray}
}

\hypersetup{
colorlinks=true,
linkcolor=black,
citecolor=black,
urlcolor=black
}

\begin{document}
\title{SafeOrchestrate: Runtime-Checked Intent-Driven Orchestration for Cyber-Physical Systems}

\author{Muhammad Ashfaq\inst{1}{\Envelope}\,\orcidlink{0000-0003-1870-7680} \and
Ahmed R. Sadik\inst{2}\orcidlink{0000-0001-8291-2211} \and
Ewan Warburton\inst{3}\orcidlink{0009-0001-1534-4872} \and
Pyry Kotilainen\inst{1}\orcidlink{0000-0002-4645-074X} \and
Niko M\"{a}kitalo\inst{1}\orcidlink{0000-0002-7994-3700} \and
Tommi Mikkonen\inst{1}\orcidlink{0000-0002-8540-9918}
}
\authorrunning{Ashfaq et al.}
\institute{Faculty of IT, University of Jyv\"askyl\"a, Jyv\"askyl\"a, 40100, Finland
\email{\{muhammad.m.ashfaq,pyry.kotilainen,niko.k.makitalo,tommi.j.mikkonen\}@jyu.fi}
\and
Honda Research Institute Europe, Offenbach, 63073, Germany
\email{ahmed.sadik@honda-ri.de}\\ \and
Independent Researcher \email{e.warburton.ew@gmail.com}}
\maketitle

\pagestyle{plain}

\begin{abstract}

Modern cyber-physical systems (CPS) integrate heterogeneous sensors, actuators, controllers, and services, composed at runtime rather than at design time.
Orchestrating such systems through natural language (NL) is attractive, but unconstrained large language model (LLM) tool use is risky because actions must respect device capabilities, runtime state, and safety constraints.
We present SafeOrchestrate, a reference architecture for runtime-checked, intent-driven CPS orchestration. Given an NL intent, it derives atomic goals, generates a runtime-checkable specification for each, grounds the specifications to device affordances, and verifies candidate actions against runtime state and safety constraints before execution.
In a Web of Things (WoT) instantiation, the architecture produces Linear Temporal Logic-style specifications from the NL intent, grounds them to Thing Description (TD) affordances exposed through the Model Context Protocol (MCP), and checks candidate actions against satisfiability modulo theories (SMT)-backed guardrails.
We evaluate this instantiation on simulated smart home, greenhouse, and industrial automation tasks against a direct LLM-to-MCP baseline.
SafeOrchestrate maintains 100\% required-action accuracy on every task and 100\% MCP Execution Similarity (MCP-ES) on all but one, where MCP-ES is 92\%. By contrast, the baseline degrades on multi-device, event-triggered, and ordered tasks, with MCP-ES falling to 13\%. Because the tasks are largely single-rule with an empty constraint set, the SMT layer's cross-rule and safety-constraint checking is demonstrated on a worked example, not quantified; within that scope, the results suggest that combining LLM-based specification generation with role-aware grounding and runtime constraint checking improves reliability for open CPS orchestration without enriched TD annotations or planning-domain expertise.


\end{abstract}

\section{Introduction}
\label{sec:introduction}

Cyber-physical systems (CPS) are increasingly assembled from independently developed constituent systems (CSs), such as sensors, actuators, controllers, and cloud services. Composed at runtime, these CSs can support systems-of-systems (SoS) behavior that no individual CS provides~\cite{sadik2023self,fortino2020internet}.
For example, a smart building can coordinate lighting, ventilation, and occupancy sensing; an agricultural setup can jointly react to temperature, humidity, and soil moisture; and an industrial cell can coordinate perception, manipulation, and material handling.
Historically, composition logic was fixed at design time: closed-world deployments rely on known CSs, fixed formats, and hard-coded rules.
Open SoS deployments, by contrast, operate in dynamic environments and must adapt to unanticipated CSs, behaviors, and events~\cite{torkjazi2024review}.
Their required behavior depends on the current deployment, available devices, and operator intent rather than on a workflow known in advance, so composition must happen at runtime.

Realizing runtime composition is difficult~\cite{ASHFAQ2026112661} because CSs are highly heterogeneous, differing in communication protocols, data schemas, interaction patterns, and operational assumptions.
To reduce this heterogeneity, different domains describe CSs with standardized, machine-interpretable capability descriptions: OpenAPI for web and software-intensive SoS~\cite{openapi}, OPC~UA and AutomationML in industrial automation~\cite{opcua,automationml}, and URDF in robotics~\cite{ros}.

The Web of Things (WoT) is one such standard, aimed at cross-domain interoperability for connected devices~\cite{wotW3C2026}.
In WoT, Thing Descriptions (TDs) describe physical or virtual Things using JSON-LD metadata and expose interaction affordances as properties, actions, and events, while protocol bindings abstract away the underlying communication mechanisms (e.g., HTTP, MQTT, or CoAP)~\cite{wotTDW3C2026}.
However, TDs mainly state what each Thing can sense or do; they do not determine how multiple Things should cooperate, how a natural-language (NL) goal should be operationalized, or whether a proposed action is safe in the current state.

Prior work approaches this gap along three lines: WoT mashup generation composes Thing affordances into higher-level behavior, typically as reusable design-time artifacts~\cite{kast2020web,korkan2021mage,korkan2024dyst}; LLM-assisted mashup synthesis turns NL intent into planning artifacts or executable flows~\cite{salama2025llm,warburton2025iot,sorici2026amihmas}; and verified LLM tool execution gates individual tool calls against pre-existing contracts~\cite{liu2026toolgate}. None provides the intervening orchestration layer: one that derives coordination logic from NL, binds it to affordances discovered at runtime, and enforces safety before any physical action.

In this paper, we present SafeOrchestrate, a reference architecture for runtime-checked, intent-driven CPS orchestration, where an \emph{intent} is an NL goal.
Instead of prompting an LLM to generate workflows directly or invoke device interfaces freely, SafeOrchestrate grounds NL-derived specifications to executable affordances and checks candidate actions before execution.
Our WoT instantiation realizes the architecture through Linear Temporal Logic (LTL)-style specification generation, TD affordances exposed as Model Context Protocol (MCP) tools and resources, and satisfiability modulo theories (SMT)-backed guardrails, orchestrating heterogeneous devices through a uniform, runtime-checked execution interface.

We investigate the following research question (RQ): 
\emph{How does grounding NL-derived specifications to MCP-exposed WoT affordances, and verifying candidate actions against runtime state before execution, affect the correctness of LLM-driven CPS orchestration, relative to direct LLM-to-MCP tool use?} 
We make the following contributions:
\begin{enumerate}[leftmargin=*,noitemsep,topsep=0pt]
\item We propose SafeOrchestrate, a reference architecture for LLM-driven CPS orchestration that integrates specification generation, role-aware grounding, and SMT-backed runtime guardrails over MCP-exposed affordances.

\item We introduce MCP Execution Similarity (MCP-ES), an alignment metric between expected and actual MCP tool-call trajectories that penalizes missing, extra, substituted, and mis-ordered calls.

\item We evaluate the WoT/MCP instantiation of SafeOrchestrate across smart home, greenhouse, and industrial automation tasks, where it matches a direct LLM-to-MCP baseline on simple tasks and substantially outperforms it on event-triggered, ordered, and multi-device tasks.

\end{enumerate}

The remainder of this paper is organized as follows. Section~\ref{sec:background-and-related-work} reviews WoT mashups, LLM-assisted workflow generation, and verified tool execution. Section~\ref{sec:proposed-method} presents the SafeOrchestrate architecture. Section~\ref{sec:implementation} describes the WoT/MCP implementation. Section~\ref{sec:evaluation} reports the evaluation, and Section~\ref{sec:conclusion} concludes with limitations and future work.
The implementation of SafeOrchestrate and all evaluation artifacts are publicly available in the accompanying repository\footnote{\url{https://github.com/ashfaq92/safeOrchestrate}}.

\section{Background and Related Work}
\label{sec:background-and-related-work}

In the WoT literature, a mashup is an application-level composition of multiple Thing affordances, such as reading a sensor property and invoking one or more actuator actions in response. WoT mashup research has long used intermediate artifacts to represent and validate such compositions. System Descriptions (SDs) encapsulate sequence-diagram-based atomic mashups as reusable affordances~\cite{kast2020web}, while DyST mines device interactions to reconstruct sequence diagrams and SDs for validation~\cite{korkan2024dyst}. A-MaGe reduces manual design effort by generating candidate atomic mashups from TDs and user options~\cite{korkan2021mage}. These approaches support reuse, generation, and validation, but remain tied to sequence diagrams, SDs, templates, rules, or user selection rather than to guarded execution over live affordances.

Several recent systems use LLMs to reduce the manual effort of creating IoT and WoT workflows from NL intent. LLM-MaGe translates user intent and TDs into Planning Domain Definition Language (PDDL) artifacts~\cite{salama2025llm}, but its TD-to-planning conversion depends on enriched annotations such as \texttt{affects} and \texttt{isAffectedBy}, limiting compatibility with ordinary TDs. IoT Whisperer generates Node-RED workflows from NL descriptions and TDs~\cite{warburton2025iot}. AmI HMAS maps Home Assistant deployments into TD-based hypermedia environments and uses LLM-driven agent roles to interpret requests, discover affordances, plan invocations, and reuse prior interactions~\cite{sorici2026amihmas}. These approaches produce intermediate planning, workflow, or hypermedia-agent representations rather than checking grounded runtime actions against MCP-exposed affordances.

MCP changes the execution model from offline artifacts to live tools and resources exposed through discoverable interfaces~\cite{mcp2026}. ToolGate~\cite{liu2026toolgate} checks Hoare-style preconditions and postconditions over symbolic state for general LLM tool use. However, this contract-gated model assumes suitable tool contracts are available---manually specified or extracted from reliable documentation---and focuses on per-tool execution.

SafeOrchestrate differs from all three: WoT mashup generation, LLM-assisted synthesis, and verified tool execution. Unlike template-, rule-, or selection-driven mashups, it derives coordination logic from NL rather than from design-time artifacts; unlike planning- and workflow-generation systems, it grounds that logic to affordances discovered at runtime over MCP, without the enriched TD annotations that systems such as LLM-MaGe require; and unlike per-tool contract verification, it checks candidate actions against the grounded current state at execution time rather than against pre-authored contracts.

\section{Proposed Method}
\label{sec:proposed-method}

\subsection{Architectural Overview}
\begin{figure}[ht]
\centering
\includegraphics[width=0.72\linewidth]{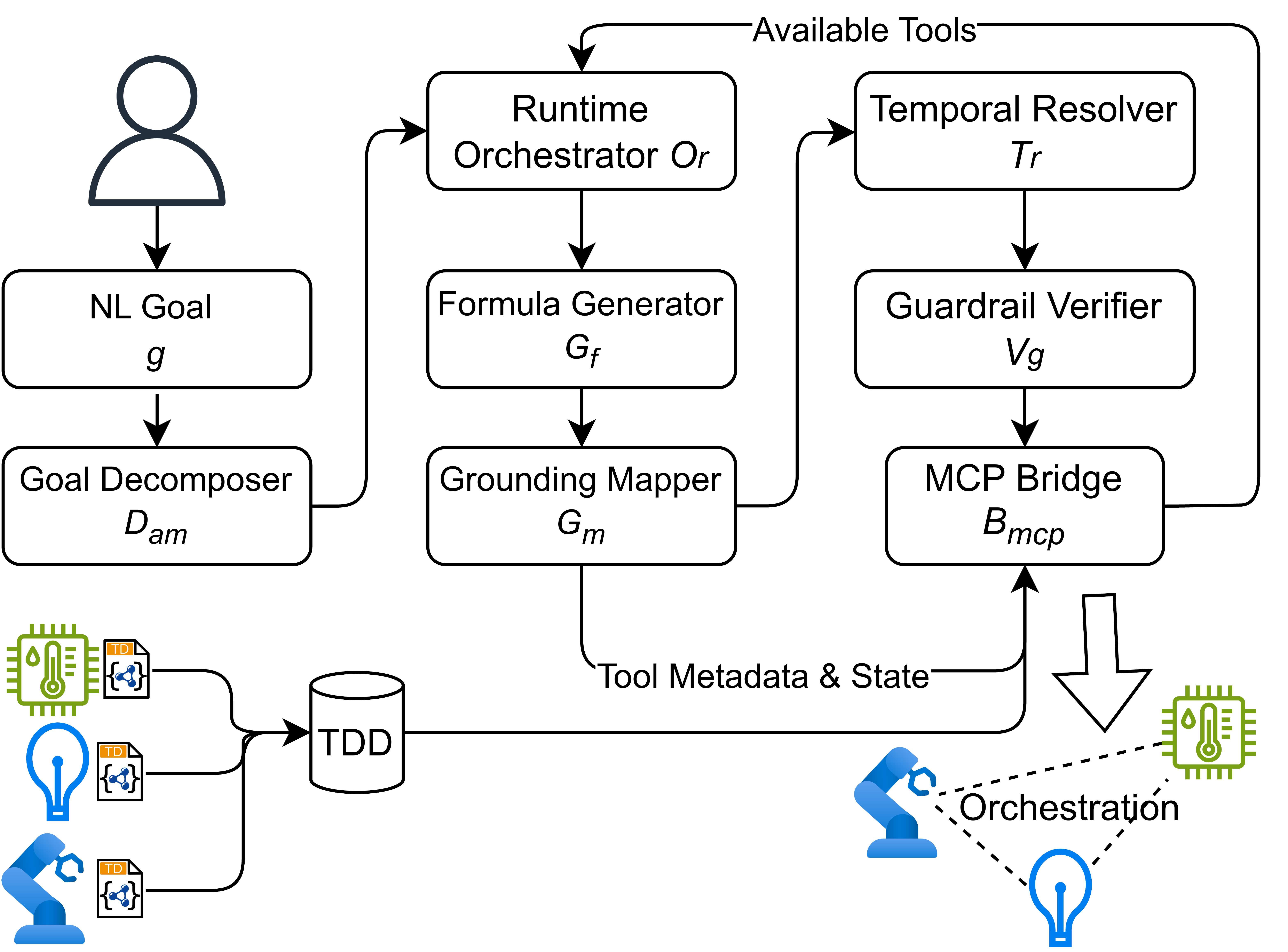}
\caption{The SafeOrchestrate reference architecture. Device Thing Descriptions are registered in a Thing Description Directory (TDD) and exposed by the MCP Bridge ($B_{\text{mcp}}$). Given an NL goal $g$, $D_{am}$ derives atomic goals; per goal, $G_f$ generates a specification, $G_m$ grounds it to affordances discovered from $B_{\text{mcp}}$ (``Available Tools''), $T_r$ produces reactive rules, and $V_g$ checks each candidate against current state (``Tool Metadata \& State'') before $O_r$ invokes it.}
\label{fig:componentOverview}
\end{figure}

SafeOrchestrate takes device capability specifications and an NL intent $g$ as input. The proposed reference architecture, shown in Fig.~\ref{fig:componentOverview}, is defined as:
$$ \mathcal{A} = \langle B_{\text{mcp}}, D_{am}, G_f, G_m, T_r, V_g, O_r \rangle $$

where $B_{\text{mcp}}$ is the \textit{MCP Bridge} (device abstraction layer); $D_{am}$ the \textit{Atomic Goal Decomposer} (compound goal $\rightarrow$ atomic mashup candidates); $G_f$ the \textit{Formula Generator} (NL $\rightarrow$ runtime-checkable specification); $G_m$ the \textit{Grounding Mapper} (symbolic $\rightarrow$ executable mapping); $T_r$ the \textit{Temporal Resolver} (formula $\rightarrow$ reactive rules); $V_g$ the \textit{Guardrail Verifier} (constraint enforcement); and $O_r$ the \textit{Runtime Orchestrator} (execution coordination).

\subsection{MCP Bridge ($B_{\text{mcp}}$)}

$B_{\text{mcp}}$ transforms device capability descriptions into MCP-compatible runtime affordances.
Actions, properties, and events become executable tools and observable resources.
At runtime, $B_{\text{mcp}}$ exposes the executable tools $\mathcal{T}_t = \{t_1, \dots, t_n\}$ and observable resources $\mathcal{R}_t = \{r_1, \dots, r_m\}$ available at time $t$.
These affordances form the substrate for grounding, guardrail checking, and execution, and they hide device protocols from subsequent components.
In the WoT instantiation, TDs are converted into MCP tools and resources; in general CPS contexts, device APIs, URDF robot descriptions, or other specifications undergo analogous transformations.

\subsection{Atomic Goal Decomposer ($D_{am}$)}

A compound NL goal often bundles several independent behaviors. $D_{am}$ splits $g$ into atomic goals $D_{am}(g) = \{am_1, am_2, \dots, am_k\}$, where each \emph{atomic goal} $am_i$ is a single, indivisible unit of intended behavior: one observed condition paired with the action it should trigger. Each $am_i$ is treated as an atomic mashup candidate (in the WoT sense, hence the subscript $am$) with input and output interactions. Inputs are observable conditions (property reads, sensor values, events, or status checks); outputs are executable effects (action invocations or property writes)~\cite{kast2020web}. This separation prevents command-like symbols from serving as state predicates and supports sequential execution of multi-sentence goals.

\subsection{Formula Generator ($G_f$)}

$G_f$ converts each atomic goal $am_i$ into a runtime-checkable specification. We represent generation as $G_f(am_i) \leadsto \phi_i$, where $\phi_i$ is the resulting LTL-style formula over symbolic propositions.

We use an LTL-style syntax because it provides a compact notation for implications, conjunctions, and reactive intent, and because existing NL-to-LTL tools can generate such specifications.

\subsection{Grounding Mapper ($G_m$)}
The generated formula $\phi$ contains symbolic proposition labels that must be mapped to executable CPS affordances. $G_m$ aligns these labels with MCP tools and resources in two stages.

\paragraph{Stage 1: Role-Aware Proposition-to-Tool Mapping.}

Symbolic proposition labels from $\phi_i$ are mapped to MCP affordances using LLM-based semantic matching over the tools and resources first discovered from $B_{\text{mcp}}$.
Although these labels are formula atoms, they are usually generated from NL phrases (e.g., \texttt{lamp\_is\_on}), so $G_m$ uses both their lexical meaning and their formula role.
Precondition propositions are grounded to observable affordances such as property reads, sensor tools, or event resources; action propositions are grounded to executable action invocations or property writes.

For a proposition $p$ and role $r \in \{\text{precondition}, \text{action}\}$, the atomic grounding map returns a subset of available MCP affordances $\Gamma_i(p,r) \subseteq \mathcal{T}_t \cup \mathcal{R}_t$, where $p \in \text{Prop}(\phi_i)$. In most cases, this subset contains a single affordance; for composite propositions such as ``all ZigBee lamps,'' it may contain several actuator tools.

\paragraph{Stage 2: Numeric Enrichment.}
For propositions containing numeric comparisons (e.g., ``temperature exceeds $26^\circ\mathrm{C}$''), $G_m$ enriches tool IDs with comparison operators, $\text{proposition} \xrightarrow{\Gamma_i} \text{tool\_id} \oplus \text{comparison}$, where $\oplus$ denotes operator attachment. For example, ``temperature exceeds 26'' maps to \texttt{get\_temperature > 26}. The fully grounded specification is $\phi_{g,i} = \Gamma_i^{*}(\phi_i)$. On an atom, $\Gamma_i^{*}(p)$ is the grounded expression for $p$: a single affordance ID, a numeric comparison over one, or---when $p$ names several affordances---their conjunction, with the role of $p$ fixed by its position in $\phi_i$. On compound formulas $\Gamma_i^{*}$ recurses without altering structure:
$$
\Gamma_i^{*}(\varphi \odot \psi) = \Gamma_i^{*}(\varphi)\,\odot\,\Gamma_i^{*}(\psi), \qquad
\Gamma_i^{*}(M\,\varphi) = M\,\Gamma_i^{*}(\varphi),
$$
for binary $\odot \in \{\land,\lor,\rightarrow,\leftrightarrow,\mathbf{U}\}$ and unary $M \in \{\lnot,\mathbf{G},\mathbf{F},\mathbf{X}\}$. Grounding leaves every logical connective and temporal operator in place and assigns the temporal operators no meaning; $T_r$ removes them (Section~\ref{subsec:temporal-resolver}), producing a current-state abstraction rather than an LTL semantics (Section~\ref{sec:threats-to-validity}). The per-stage grounding rules, and the constructs the prototype does not resolve (e.g., \emph{until}, bounded eventualities), are detailed in the supplementary appendix of the accompanying repository.

\subsection{Temporal Resolver ($T_r$)}
\label{subsec:temporal-resolver}

The grounded specification $\phi_g$ is transformed into \emph{reactive precondition-action rules} evaluated over the current runtime state $s_t$:

$$
T_r(\phi_g) =
\{(pre_1, act_1), (pre_2, act_2), \dots, (pre_k, act_k)\}
$$

where the temporal operators are removed and each pair $(pre_i, act_i)$ is then extracted from the implication and biconditional patterns in the resulting propositional formula: $(A \rightarrow B)$ yields $(A, B)$, and $(A \leftrightarrow B)$ yields $(A, B)$ and $(B, A)$.

\subsection{Guardrail Verifier ($V_g$)}
\label{subsec:guardrail}

Before execution, $V_g$ evaluates candidate MCP tools against the grounded specification and current state, enforcing system-level constraints in two phases.

\paragraph{Phase 1: Precondition Evaluation}
For each precondition-action pair $(pre_i, act_i) \in T_r(\phi_g)$, $V_g$ evaluates the precondition against the current state, $\text{eval}(pre_i, s_t) \in \{\text{true}, \text{false}\}$. If $\text{eval}(pre_i, s_t) = \text{false}$, action $act_i$ is rejected before SMT checking.

\paragraph{Phase 2: SMT-Based Constraint Check.}

For actions with satisfied preconditions that come from a single-action rule ($O_r$ skips this phase for multi-action consequents; Section~\ref{subsec:orchestrator}), $V_g$ checks whether the candidate action is compatible with the grounded current-state constraints, $V_g(\phi_g, s_t, t_i) \in \{\text{true}, \text{false}\}$, where $\phi_g$ is the grounded specification, $s_t$ the current state, and $t_i \in \mathcal{T}_t$ the candidate MCP tool. Let $\psi_g$ denote the propositional abstraction of $\phi_g$ obtained by removing its temporal operators, and let $C$ denote optional deployment safety constraints when available. $V_g$ checks satisfiability of the propositional abstraction:

$$
\text{SAT}(\psi_g \land C \land \widehat{s_t} \land (t_i = \text{true}))
$$

where $\widehat{s_t}$ encodes the observed runtime values as Boolean or numeric assignments. If satisfiable, execution is permitted; otherwise, the action is blocked. Because the query ranges over the entire specification $\psi_g$, not just the rule that selected $t_i$, it blocks an action that satisfies its precondition yet conflicts with another rule under the current state (for example, the reverse direction of a biconditional). This consistency check is active even when $C$ is empty, reducing to the Phase~1 condition only for an isolated implication. The optional set $C$ adds deployment safety constraints, such as an emergency-stop rule; in our evaluation, $C$ is empty.

\subsection{Runtime Orchestrator ($O_r$)}
\label{subsec:orchestrator}
\begin{figure}[t]
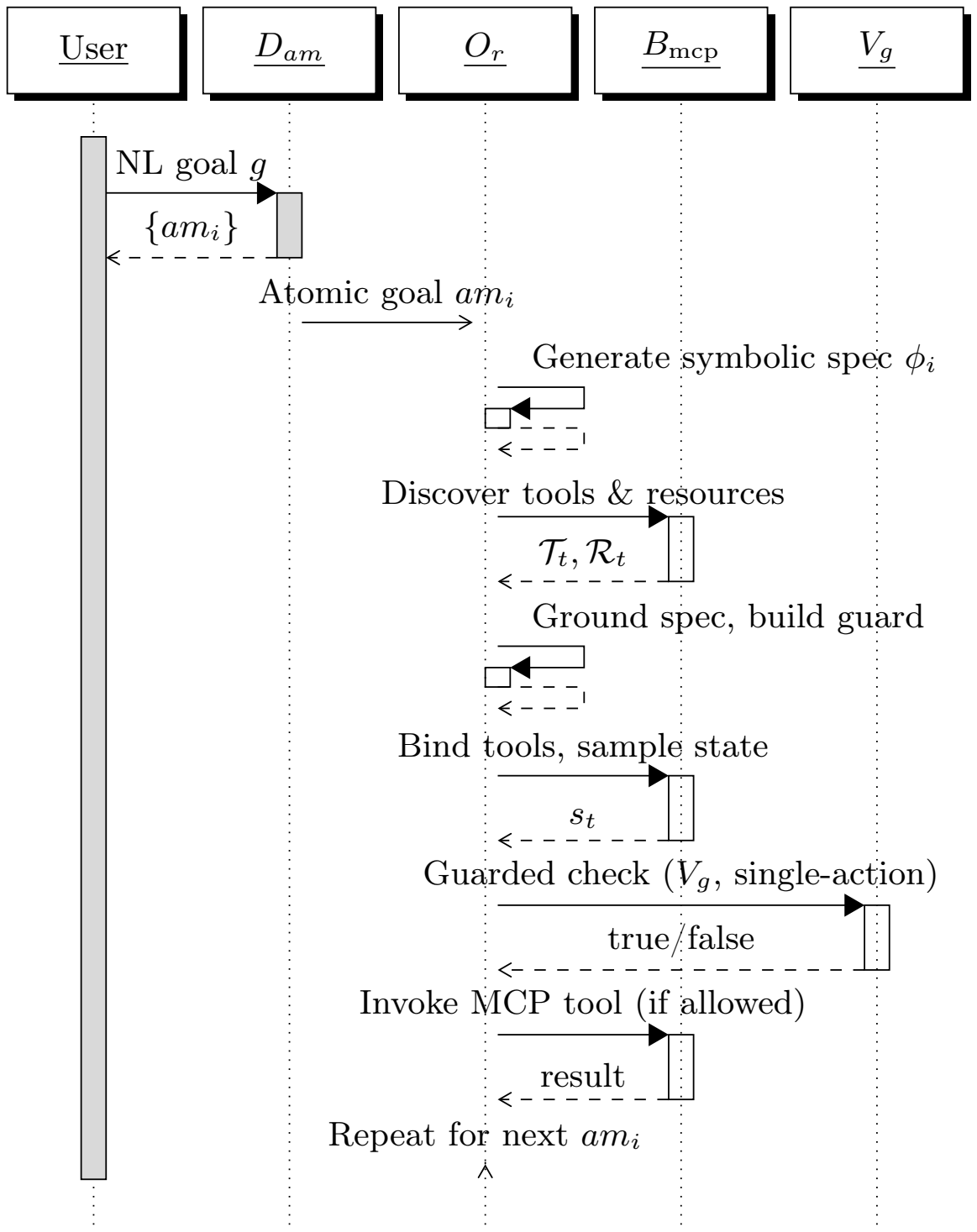

\centering
 \scalebox{0.85}{%
 \begin{sequencediagram}
\newthread{user}{User}
\newinst{decomposer}{$D_{am}$}
\newinst{orchestrator}{$O_r$}
\newinst{bridge}{$B_{\text{mcp}}$}
\newinst{verifier}{$V_g$}

\begin{call}{user}{NL goal $g$}{decomposer}{$\{am_i\}$}
\end{call}

\mess{decomposer}{Atomic goal $am_i$}{orchestrator}

\begin{callself}{orchestrator}{Generate symbolic spec $\phi_i$}{}
\end{callself}

\begin{call}{orchestrator}{Discover tools \& resources}{bridge}{$\mathcal{T}_t, \mathcal{R}_t$}
\end{call}

\begin{callself}{orchestrator}{Ground spec, build guard}{}
\end{callself}

\begin{call}{orchestrator}{Bind tools, sample state}{bridge}{$s_t$}
\end{call}

\begin{call}{orchestrator}{Guarded check ($V_g$, single-action)}{verifier}{true/false}
\end{call}

\begin{call}{orchestrator}{Invoke MCP tool (if allowed)}{bridge}{result}
\end{call}

\mess{orchestrator}{Repeat for next $am_i$}{orchestrator}
\end{sequencediagram}
}
\caption{Execution sequence for one atomic goal; goals are processed in turn. Rules with multiple consequent actions bypass the $V_g$ check.}
\label{fig:safe_runtime_loop}
\end{figure}

$O_r$ is the architecture's reactive component. The prototype does not maintain an internal recurring loop: the decomposer's atomic goals $am_1,\dots,am_k$ are evaluated once each, sequentially in textual order, and a fresh evaluation of a goal's resolved rules $T_r(\phi_g)$ occurs only on a new invocation (for example, the next event-triggered run). One evaluation has three steps (Fig.~\ref{fig:safe_runtime_loop}):

\begin{enumerate}[noitemsep,topsep=0pt]
\item \textbf{State sampling.} Read the observable affordances referenced by $\phi_g$---properties, sensors, and event resources---once, into the snapshot $s_t$. The snapshot is not refreshed between actions of the same goal.
\item \textbf{Candidate selection.} For each pair $(pre_i, act_i) \in T_r(\phi_g)$, evaluate $\text{eval}(pre_i, s_t)$; a satisfied precondition means $act_i$ is \emph{enabled}, and it joins $\mathcal{T}_{\text{candidates}}$.
\item \textbf{Check and execute.} A candidate whose triggering rule has a single consequent action is admitted only if $V_g(\phi_g, s_t, t_i)$ also holds---an SMT test of \emph{consistency} with $\psi_g \land C$ under $s_t$, kept separate from the precondition so that an action can be blocked for conflicting with another rule, not only for a missing trigger (Section~\ref{subsec:guardrail}). A rule whose consequent bundles several actions is currently admitted on the satisfied precondition alone (Section~\ref{subsec:discussion}). If no candidate is admitted, $\mathcal{T}_e = \emptyset$ and $O_r$ takes no action for that goal: abstention is the deliberate response when no candidate is both enabled and admitted, with no fallback or default action. Failures in the earlier LLM stages are handled likewise (Section~\ref{sec:implementation}).
\end{enumerate}

All checks run against the single snapshot $s_t$, so $O_r$ establishes that an invocation is \emph{authorized} under the sampled state---not that execution succeeds or that a physical outcome follows. It does not monitor or recover from failures during actuation itself (device errors, timeouts, unsafe transients), which would need closed-loop monitoring, and it does not establish trace-level temporal correctness (Section~\ref{sec:threats-to-validity}). Processing each goal against its own single snapshot keeps that goal's evaluation stable and well defined; the trade-off is that concurrent interleavings are not modeled and the snapshot can go stale across a multi-action sequence.

$O_r$ maps $(\phi_g, s_t) \rightarrow \mathcal{T}_e \subseteq \mathcal{T}_t$, the set of tools authorized for execution in the evaluation:

$$
\mathcal{T}_e = \{\, t \in \mathcal{T}_t \;\mid\; \exists\, (pre, act) \in T_r(\phi_g) :\; t \in act \;\land\; \text{eval}(pre, s_t) \;\land\; \big( |act| = 1 \Rightarrow V_g(\phi_g, s_t, t) \big) \,\}
$$

where $t \in act$ means $t$ is one of the consequent's actions and $|act|$ their number: a single-action rule requires the guardrail check, whereas for a multi-action consequent the guard condition is vacuously true and admission rests on the precondition alone.

\paragraph{Guarantee.}
By construction, $O_r$ never authorizes an action whose precondition is unmet, and every \emph{individually checked} $t \in \mathcal{T}_e$---one whose triggering rule has a single consequent action---is also constraint-consistent with $\psi_g \land C$ under $\widehat{s_t}$. Actions from multi-action consequents are admitted on their precondition alone, and the guarantee is over authorization under the sampled state, not physical effect ($C$ is optional; Section~\ref{subsec:discussion} discusses its scope in our evaluation). Cases~3 and~4 (Section~\ref{sec:demo}) show the blocking behavior: when $\widehat{s_t}$ falsifies a constraint in $\psi_g \land C$, the query is unsatisfiable for all candidates and $\mathcal{T}_e = \emptyset$.

\subsection{Demonstrative Example: Industrial Conveyor Control}
\label{sec:demo}
We illustrate the architecture using an industrial conveyor scenario. Let $g$ be: ``If the robot arm has an item in its hand, the belt should move forward.'' $G_f$ produces:
\[
\phi = (\textit{item-in-hand-robotArm} \rightarrow \textit{move-forward-belt}).
\]

$G_m$ grounds the specification via role-aware mapping:
\[
\begin{aligned}
\textit{item-in-hand-robotArm} &\mapsto \texttt{get\_holdingObject}, \\
\textit{move-forward-belt} &\mapsto \texttt{startBeltForward}.
\end{aligned}
\]

The lifted formula is:
\[
\phi_g = (\texttt{get\_holdingObject} \rightarrow \texttt{startBeltForward}),
\]
from which $T_r$ extracts the precondition-action pair:
\[
T_r(\phi_g) =
\{(\texttt{get\_holdingObject}, \texttt{startBeltForward})\}.
\]

\paragraph{Cases 1--2: Precondition Gating.}
If $s_t(\texttt{get\_holdingObject}) = \text{true}$, the precondition holds and the verifier admits the action, $O_r(\phi_g, s_t) = \{\texttt{startBeltForward}\}$; if it is \text{false}, no action is triggered and $O_r(\phi_g, s_t) = \emptyset$.

\paragraph{Case 3: Cross-Rule Consistency.}
Add the rule $\texttt{get\_obstacleDetected} \rightarrow \neg\,\texttt{startBeltForward}$. If both $s_t(\texttt{get\_holdingObject})$ and $s_t(\texttt{get\_obstacleDetected})$ are \text{true}, then $\text{SAT}(\psi_g \land \widehat{s_t} \land (\texttt{startBeltForward} = \text{true}))$ is unsatisfiable---the candidate conflicts with the second rule---so $O_r(\phi_g, s_t) = \emptyset$.

\paragraph{Case 4: Safety Constraint Violation Prevention.}
Let $C = \neg\,\texttt{get\_emergencyStop}$. If $s_t(\texttt{get\_emergencyStop}) = \text{true}$, then $\text{SAT}(\psi_g \land C \land \widehat{s_t} \land (\texttt{startBeltForward} = \text{true}))$ fails, yielding $O_r(\phi_g, s_t) = \emptyset$.

\section{Implementation and Design Choices}
\label{sec:implementation}

\paragraph{MCP Bridge.}
We use \texttt{WoT-MCP}\footnote{\url{https://github.com/macc-n/wot-mcp/}} as $B_{\text{mcp}}$, which translates TDs into MCP tools and resources using two strategies: (i) \textit{Explicit}, creating separate tools for property access and actions while exposing events as resources, and (ii) \textit{Generic}, providing unified tools for discovery, property access, and invocation.

\paragraph{Specification Generation.}
We adopt \texttt{nl2spec}~\cite{cosler2023nl2spec}, which most reliably produced parseable outputs in preliminary trials. It is exposed via an API for $G_f$. We extend its prompt to enforce role-consistent propositions, mapping observable conditions to state propositions and effects to action propositions, while keeping few-shot examples domain-agnostic.

\paragraph{Grounding Mapper.}
$G_m$ performs LLM-based mapping and validation, producing atomic mappings $\Gamma_i(\cdot)$ and applying $\Gamma_i^{*}$ to obtain grounded formulas. Action propositions map to actuator commands (e.g., \texttt{turnOn\_lamp}), while state propositions map to sensor resources (e.g., \texttt{get\_isOn\_lamp}). Numeric constraints (e.g., \texttt{get\_temperature > 26}) are extracted using regex-based detection of thresholds and comparison keywords.

\paragraph{Temporal Resolution and Precondition Extraction.}
$T_r$ removes temporal operators ($\mathbf{G}$, $\mathbf{F}$, $\mathbf{X}$, $\mathbf{U}$) to obtain propositional formulas evaluated over the current state $s_t$. For example, $\mathbf{G}(A \rightarrow B)$ reduces to $(A \rightarrow B)$. Operators are dropped rather than semantically encoded; the appendix tabulates the per-operator handling and the constructs for which the prototype provides no temporal guarantee. The resulting formulas are parsed using a Lark-based grammar into an AST, traversed by an \texttt{ImplicationExtractor} to identify implication and biconditional patterns.

\paragraph{Guardrail Verifier.}
$V_g$ uses the Z3 SMT solver\footnote{\url{https://github.com/z3prover/z3}} for constraint checking. Tool IDs are aliased to variables (\texttt{v0}, \texttt{v1}, \ldots) and typed as \texttt{Bool} or \texttt{Real} depending on usage in numeric comparisons. Expressions are translated into Z3 calls via a shunting-yard-inspired algorithm, e.g., $A \land B \rightarrow C$ becomes \texttt{Implies(And(A, B), C)}.

\paragraph{Robustness to malformed LLM outputs.}
Only $G_f$, $G_m$, and argument synthesis invoke an LLM; $B_{\text{mcp}}$ and $D_{am}$ are deterministic (schema transformation; rule-based splitting, with a non-decomposable prompt passed through). Each LLM stage has a deterministic check or a hard failure behind it. $G_f$ output is normalized; an empty result aborts the goal; there is no syntax preflight, but the execution parser fails closed, so an unparseable formula produces no action. $G_m$ output must parse as JSON (invalid JSON aborts the invocation); a role-aware well-formedness filter then drops pairs whose antecedent is not observable or whose consequent is not executable. $G_m$ is not guaranteed to ground every atom and there is no completeness check: an unmapped atom is dropped downstream---by that filter or the execution-time tool lookup---so the sub-goal produces no action, logged but with no user-facing error; recovery, where enabled, is interactive revision of the intent. Synthesized arguments are merged with schema-derived defaults, unvalidated. Malformed formulas, unmapped affordances, and bad arguments thus do not produce executed actions; semantically wrong but well-formed formulas, mappings, and argument values are the residual risk (Section~\ref{sec:threats-to-validity}). The supplementary appendix tabulates these checks.

\section{Evaluation}
\label{sec:evaluation}

We evaluate the WoT instantiation of SafeOrchestrate with respect to the research question in Section~\ref{sec:introduction}. Full task prompts and execution logs are available in the accompanying repository. 
We consider eight CPS orchestration tasks drawn from prior work to assess system performance under diverse control patterns. Table~\ref{tab:evaluation-scenarios} summarizes their descriptions and primary orchestration capabilities.

\begin{table}[ht]
\centering
\scriptsize
\begin{tabular}{p{0.30\linewidth}p{0.40\linewidth}p{0.28\linewidth}}
\hline
Scenario & Description & Primary capability \\
\hline
Presence Control~\cite{salama2025llm} & Automate lighting from room occupancy. & Property-triggered condition-action \\
AC Control~\cite{salama2025llm} & Maintain room temperature via threshold-based switching. & Bidirectional threshold control \\
Robot Workflow~\cite{salama2025llm} & Coordinate perception, manipulation, and actuation steps. & Sequential multi-stage control \\
Greenhouse Automation~\cite{salama2025llm} & Maintain plant conditions via coupled climate rules. & Multi-condition coordination \\
Washing Machine Event~\cite{warburton2025iot} & Trigger LED blinking after cycle completion. & Event-triggered action \\
Doorbell Workflow~\cite{warburton2025iot} & Temporarily adjust audio, notify user, and restore state. & Ordered event-driven workflow \\
Alarm Ringing~\cite{warburton2025iot} & Maintain temperature for fixed duration after alarm trigger. & Timed event execution \\
Coming/Leaving Home~\cite{kast2020web} & Control multiple devices for home arrival/departure. & Multi-device coordination \\
\hline
\end{tabular}
\caption{Evaluation tasks grouped by orchestration capability.}
\label{tab:evaluation-scenarios}
\end{table}

\subsection{Experimental Setup}

\paragraph{System configuration.}
We compare SafeOrchestrate against a VanillaOrchestrate baseline that directly maps atomic goals to MCP tool calls, bypassing runtime verification. To prevent runaway behavior under non-convergence, the baseline is capped at three reactive tool-calling steps. SafeOrchestrate executes the full pipeline after decomposition, including role-aware grounding, constraint formulation, MCP invocation, and guardrail verification. For fairness, we disable human-in-the-loop intervention and iterative refinement for SafeOrchestrate.

\paragraph{Model and execution settings.}
All components use GPT-4.1 except \texttt{nl2spec}, which runs with GPT-3.5 Turbo to match prior work~\cite{cosler2023nl2spec}. Temperature is fixed to 0. Due to multiple LLM calls per run and provider-side nondeterminism, results vary across executions. We therefore report averages over ten runs per atomic goal.

\paragraph{Evaluation protocol.}
We reconstruct oracle action sequences from execution logs and compare them against observed tool-call traces. This enables us to compute required-action accuracy and MCP-ES (Section~\ref{sec:mcp-es}). For event-driven tasks, we reset the event buffer before each run and inject the trigger exactly once in event-present settings. We additionally include event-absent controls to verify that systems do not act without triggers; these serve as diagnostic checks rather than primary evaluation conditions.

\subsection{Evaluation Metrics}
\label{sec:mcp-es}

We use two metrics that capture complementary aspects of orchestration quality.

\paragraph{Required-action accuracy.}
Following prior work~\cite{salama2025llm}, we measure whether all oracle actions are correctly executed at least once. This metric captures functional correctness but is insensitive to ordering and extraneous calls.

\paragraph{MCP Execution Similarity (MCP-ES).}
To capture full trace fidelity, we propose MCP-ES as the normalized Levenshtein similarity between expected and executed MCP call sequences $E$ and $A$:
\[
\text{MCP-ES} = 1 - \frac{D_{\text{edit}}(E, A)}{\max(|E|, |A|)}.
\]

MCP-ES penalizes insertions, deletions, substitutions, and ordering errors. Before computing distance, we remove read-only state queries (e.g., \texttt{get\_*}) and internal guardrail inspection calls that do not affect system state or execution order. All state-changing actions (tool invocations and property writes) are retained. The same filtering is applied to both SafeOrchestrate and VanillaOrchestrate to ensure comparability. For composite instructions, oracle actions are expanded to all required actuations, while unintended additional actuations are counted as insertions.

\subsection{Results}

\begin{table}[ht]
\centering
\scriptsize
\setlength{\tabcolsep}{5pt}
\begin{tabular}{lrrrr}
\hline
 & \multicolumn{2}{c}{VanillaOrchestrate} & \multicolumn{2}{c}{SafeOrchestrate} \\
\cline{2-3}\cline{4-5}
Task & Acc.\ & MCP-ES & Acc.\ & MCP-ES \\
\hline
Presence Control      & 100 & 100 & 100 & 100 \\
AC Control            & 100 & 100 & 100 & 100 \\
Robot Workflow        & 80  & 75  & 100 & 100 \\
Greenhouse Automation & 83  & 87  & 100 & 100 \\
Washing Machine Event & 50  & 50  & 100 & 100 \\
Doorbell Workflow     & 17  & 13  & 100 & 100 \\
Alarm Ringing         & 25  & 25  & 100 & 100 \\
Coming/Leaving Home   & 80  & 17  & 100 & 92  \\
\hline
\end{tabular}
\caption{Required-action accuracy and MCP-ES (\%, mean of ten runs per atomic goal). SafeOrchestrate reaches 100\% throughout except Coming/Leaving Home MCP-ES (92\%, one extra non-oracle lighting action); VanillaOrchestrate degrades on event-triggered, ordered, and multi-device tasks.}
\label{tab:results-summary}
\end{table}

Table~\ref{tab:results-summary} reports per-task required-action accuracy and MCP-ES for both systems.

Both approaches perform similarly on simple property-triggered tasks, but VanillaOrchestrate degrades as task complexity, event dependence, and device count increase---to 80\%/75\% (accuracy/MCP-ES) in Robot Workflow and 83\%/87\% in Greenhouse Automation.

The gap widens substantially in event-triggered and multi-device settings: 17\%/13\% in Doorbell Workflow, and in Coming/Leaving Home 80\% accuracy but only 17\% MCP-ES, indicating severe ordering and coordination failures.

SafeOrchestrate remains stable across all tasks, achieving 100\% required-action accuracy throughout. It also attains 100\% MCP-ES in all tasks except Coming/Leaving Home, where it achieves 92\% MCP-ES due to a single additional non-oracle lighting action.

The logs for each atomic goal reveal distinct baseline failure patterns, which we annotate using the taxonomy in Table~\ref{tab:failure-taxonomy}. In Robot Workflow, failures concentrate at the final placement step. In Greenhouse Automation, errors accumulate across multiple environmental constraints. In Coming/Leaving Home, required-action accuracy remains relatively high, but MCP-ES drops sharply due to ordering violations and extraneous actions.

\begin{table}[ht]
\centering
\scriptsize
\begin{tabular}{p{0.24\linewidth}p{0.43\linewidth}p{0.34\linewidth}}
\hline
Failure type & Definition & Observed examples \\
\hline
Skipped precondition & Executes an action without observing the required state or event. & Negative-control event runs; inactive coming/leaving branch execution \\
Wrong tool & Selects an affordance with the wrong role or semantics. & Confusing state-query and actuator affordances in climate-control runs \\
Missing action & Omits one or more oracle actions. & Doorbell Workflow missing volume reduction or restoration \\
Wrong order & Executes required actions in a sequence that breaks the workflow semantics. & Ordered robot and doorbell workflows \\
Wrong argument & Uses a correct action with invalid or semantically wrong parameters. & Speaker volume parameter mismatch in VanillaOrchestrate runs \\
Unsafe extra action & Invokes an additional action not permitted by the active trigger/state. & Leaving-home branch invoked during coming-home request; extra light-off action \\
\hline
\end{tabular}
\caption{Failure taxonomy used when interpreting execution logs.}
\label{tab:failure-taxonomy}
\end{table}

\subsection{Discussion}
\label{subsec:discussion}

\paragraph{Baseline failure modes.}
Purely LLM-driven orchestration is adequate for simple tasks but unreliable under
event-driven control, ordered execution, and multi-device coordination---exactly
where explicit grounding and runtime checking help most. The taxonomy
(Table~\ref{tab:failure-taxonomy}) highlights systematic VanillaOrchestrate
weaknesses: missing precondition checks, premature termination, invalid arguments,
state-query/action confusion (\texttt{turnOnAC} vs.\ \texttt{getIsOnAC}),
and---in event-triggered settings like Coming/Leaving Home---trigger indiscipline,
where actions fire regardless of the active event. Role-aware grounding and
guardrail filtering mitigate these. SafeOrchestrate reaches full required-action
accuracy without iteration on these tasks, consistent with structured grounding
plus runtime guardrails helping beyond prompt-level planning alone.

\paragraph{Scope of the safety claims.}
We scope the safety interpretation of these results accordingly. In our
evaluation, the deployment-constraint set $C$ is empty, so guardrail enforcement
primarily ensures precondition satisfaction and trigger discipline rather than
explicit safety policies. Phase~2 can resolve cross-rule conflicts at
runtime, but the reported evaluation does not exercise any---the behavior is shown
only in the worked example (Section~\ref{sec:demo})---and Phase~2 is in any case
bypassed for rules whose consequent bundles multiple actions (e.g., Coming/Leaving
Home), which are admitted on the satisfied precondition alone. Enforcement is also
entirely pre-execution: the
orchestrator does not inspect action results, re-check state after acting, or
recover from actuation faults, so hazards arising during or after physical
execution are outside the current design. This reflects a runtime verification
model rather than full offline assurance.

\paragraph{Limitations of the metrics.}
Required-action accuracy is insensitive to traces; MCP-ES captures trajectory
fidelity but penalizes benign non-oracle behavior (e.g., extra device
deactivation in leaving-home). We read MCP-ES as a trace-conformance metric, not a
semantic-correctness measure: a system can score high on accuracy yet still emit
extraneous actions.
\subsection{Threats to Validity}
\label{sec:threats-to-validity}

\paragraph{Internal Validity.}
Stochastic execution and the lack of component-wise ablation threaten internal validity: ten repetitions per task do not guarantee exhaustive coverage of stochastic execution paths.

\paragraph{External Validity.}
The evaluation suite covers condition-action rules, event-triggered actions, ordered event reactions, and multi-device coordination. It excludes long-running cyclic workflows and dynamic branching where later actions depend on intermediate execution outputs~\cite{kast2020web}. Broader evaluation in robotics and industrial CPS domains is therefore needed.
An empirical head-to-head with LLM-MaGe or IoT Whisperer is a comparability limitation, not just future work: they emit different artifacts (PDDL plans, Node-RED flows) with no shared execution-trace metric, and LLM-MaGe further requires enriched TD annotations absent from our tasks; Sections~\ref{sec:background-and-related-work} and~\ref{subsec:discussion} give the conceptual differences. We therefore restrict our empirical claims to the VanillaOrchestrate baseline and treat LLM-MaGe and IoT Whisperer as conceptual comparisons only. Broader model coverage beyond GPT-4.1 and additional baselines are further gated by the current manual reconstruction of oracle traces and MCP-ES scores, which the automated trajectory scoring planned in Section~\ref{sec:conclusion} would relieve. Finally, the evaluation focuses on required-action accuracy and MCP-ES, leaving latency, runtime efficiency, and scalability unevaluated.

\paragraph{Construct Validity: Semantic Grounding.}
SafeOrchestrate relies on semantic alignment between NL inputs, generated proposition labels, TD affordances, and MCP tool names. MCP gives a uniform execution interface but enforces no shared semantic model across these, so ambiguity or inconsistency can cause grounding errors. Role-aware grounding, parser-based normalization, and runtime guardrails mitigate this; mission-critical deployments would still need human review, stronger domain models, or deployment-specific validation.

\paragraph{Construct Validity: Runtime Checking, Concurrency, and Interleavings.}
$V_g$ performs reactive per-action constraint checking in the current state, not temporal model checking over traces---low-latency decisions without long-horizon guarantees, between online trace monitors (e.g., RV-Monitor~\cite{luo2014rv}) and offline model checking (e.g., TLA+~\cite{lamport2003tlaplus}). It also does not reason about concurrent interleavings: within an atomic goal, all precondition-action pairs are checked against one state snapshot (plus a one-step hypothetical for the action under test), not re-sampled per pair or searched over execution orders, so an action inadmissible under that snapshot is not retried in the pass. Ordering is handled by decomposition---$D_{am}$ emits single condition-action units run as a sequence of atomic goals that each re-sample state---and a subsequent external invocation (next event, next poll) re-checks a blocked action against a fresh snapshot, so a rejection is not permanent. Goals with runtime data dependencies between pairs, external state mutation mid-evaluation, and preemption of an ongoing sequence~\cite{kast2020web} are out of scope. Re-sampling state before each pair would remove the within-goal case and is a small, self-contained refinement; it, and integrating per-action checks with trace-level and offline verification, are future work.

\paragraph{Construct Validity: Temporal Specification Limitations.}
For timed tasks, the NL-to-specification stage may emit unsupported or partially formed temporal expressions (bounded eventualities, incomplete \textit{until}). Duration-based requirements are then handled only when expressible as parameters of executable affordances, not when they require trace-level temporal reasoning.

\section{Conclusion and Future Work}
\label{sec:conclusion}

This paper presents SafeOrchestrate, a runtime-checked framework for LLM-driven CPS control that combines MCP-based tool execution with guardrails. The results show improved reliability over purely LLM-driven orchestration, especially on event-triggered, ordered, and multi-device tasks. The evaluation exercises precondition satisfaction and trigger discipline; because its tasks are largely single-rule with an empty constraint set $C$, the SMT layer's cross-rule and safety-constraint enforcement is shown only on the worked example (Section~\ref{sec:demo}) and remains to be assessed at scale.

Future work will make generated orchestrations reusable through structured workflow representations, such as SDs~\cite{kast2020web} or graph-based models, building on prior work on IoT workflow serialization~\cite{warburton2025iot} and potentially grammar-based generation~\cite{mundler2025constrained}.
We also aim to decouple the orchestration logic from the MCP-specific backend by supporting direct TD consumption, where actions, properties, and events are discovered from TDs and invoked through their native protocol bindings.
Finally, we plan component-wise ablation to isolate the contribution of role-aware grounding versus guardrail checking, and automated trajectory scoring to replace manual log inspection. We also plan to evaluate safety-constraint enforcement and multi-rule conflicts across deployments at scale. Broader evaluation across long-horizon, concurrent, and cyclic CPS scenarios, additional LLMs and small language models, and larger deployments remains necessary to assess generality and scalability.

\begin{credits}
\subsubsection{\ackname}
The authors thank Muhammad Shahzad Khan, Ege Korkan, Christophe Fran\c{c}ois, and Abhishek Kumar for feedback on the architecture and evaluation design.

\subsubsection{Use of Generative AI.}
Throughout the preparation of this article, the authors used OpenAI ChatGPT, OpenAI Codex, Anthropic Claude, and Overleaf's AI writing assistant for brainstorming, for revising prose and improving grammar, clarity, and concision, and as coding assistants during the implementation and review of the accompanying prototype.
The authors independently designed the study, executed the experiments, and verified the implementation, execution logs, analysis, figures, and reported metrics. All AI-assisted text and code were reviewed, edited, and verified by the authors, and all cited literature was checked against the original sources. AI assistance in preparing the article was not used to fabricate, falsify, or manipulate research data, figures, or results. The experimental use of language models is described separately in Sections~\ref{sec:implementation} and~\ref{sec:evaluation}. The authors take full responsibility for the content of the article.

\end{credits}

\bibliographystyle{template/splncs04}
\bibliography{references}

\end{document}